\documentclass[12pt]{article}
\usepackage{amssymb}
\usepackage{color}
\usepackage{graphicx}
\newcommand {\Mybox}{Q.E.D.}

\def\x{{x}}
\def\y{{y}}
\newcommand {\E}{\mbox{E}}

\newcommand{\N}{\mathbb{N}}
\newcommand{\Z}{\mathbb{Z}}

\newcommand{\leadingones}{\mbox{\sc LeadingOnes} }
\newcommand{\OM}{\mbox{\sc OneMax} }

\newtheorem{theorem}{Theorem}

\newtheorem{corollary}{Corollary}
\newtheorem{proposition}{Proposition}
\newtheorem{definition}{Definition}

\begin{document}

\pagestyle{empty}

\title{HITTING TIMES
OF LOCAL AND GLOBAL OPTIMA IN GENETIC ALGORITHMS WITH VERY HIGH
SELECTION PRESSURE\thanks{Omsk Branch of Sobolev Institute of
Mathematics SB RAS, Omsk State University n.a.~F.M.~Dostoevsky,
eremeev@ofim.oscsbras.ru}}

\author{A.V. Eremeev}

\maketitle

\begin{abstract}
The paper is devoted to upper bounds on the expected first hitting
times of the sets of local or global optima for non-elitist
genetic algorithms with very high selection pressure. The results
of this paper extend the range of situations where the upper
bounds on the expected runtime are known for genetic algorithms
and apply, in particular, to the Canonical Genetic Algorithm. The
obtained bounds do not require the probability of
fitness-decreasing mutation to be bounded by a constant less than
one.


\end{abstract}

\section{INTRODUCTION}

\noindent The genetic algorithms~(GAs) are randomized heuristic
algorithms employing a population of tentative solutions
(individuals), which is iteratively updated by means of selection,
mutation and crossover operators, thus simulating an evolutionary
type of search for optimal or near-optimal solutions. Different
modifications of GAs are widely used in areas of operations
research and artificial intelligence.  A wider class of {\em
evolutionary algorithms} (EAs), has a more flexible outline,
possibly neglecting the crossover operator and admitting a
population which consists of a single individual. Two major types
of evolutionary algorithm outline are now well-known: the {\em
elitist} EAs keep a certain number of ``most promising''
individuals from the previous iteration, while the {\em
non-elitist} EAs compute all individuals of a new population
independently using the same randomized procedure.

The theoretical analysis of GAs has been subject to increasing
interest over the last two decades and several different
approaches have been developed. A significant progress in
understanding of non-elitist GAs was made in~\cite{Vose} by means
of dynamical systems. However most of the findings in~\cite{Vose}
apply to the infinite population case, so it is not clear how
these results can be used to estimate the runtime of GAs, i.e. the
expected number of individuals computed and evaluated until the
optimum is found for the first time. A theoretical possibility of
constructing GAs that provably optimize an objective function with
high probability in polynomial time was shown in~\cite{V00} using
rapidly mixing Markov chains. However~\cite{V00} provides only a
very simple artificial example where this approach is applicable.
The drift analysis was first adapted to studying elitist EAs
in~\cite{HY01} and further extended in~\cite{DL16,L10} to
non-elitist EAs without a crossover.

A series of works has attempted to show that the use of crossover
operator in GAs and other evolutionary algorithms can reduce their
runtime (see e.g.~\cite{JW02,KotzingSudholtTheile2011,LY11,s12})
but most of the positive results apply to families of problem
instances with specific structure. At the same time~\cite{OW14}
showed that a well-known non-elitist GA with proportional
selection operator is inefficient on one of the most simple
benchmark functions~$\OM$, even when the crossover is used.
In~\cite{MS15}, a general runtime result is proposed for a class
of convex search algorithms, including many non-elitist GAs
without mutation, on the so-called concave fitness landscapes (a
discrete-space counterpart of a concave maximization problem). As
a corollary, for another well-known benchmark
function~$\leadingones$, it is shown that the convex search
algorithm has $O(n \log n)$ runtime, which means it is faster than
all EAs using only mutation~\cite{MS15}. Upper bounds obtained for
the runtime of GAs with crossover in~\cite{CDEL14} match (up to a
constant factor) the analogous upper bounds known for
mutation-only GAs~\cite{DL16}. In~\cite{Er12}, sufficient
conditions are found under which a non-elitist GA with tournament
selection first visits a local optimum of a pseudo-Boolean
function in polynomially bounded time on average. The bounds
from~\cite{Er12} indicate that if a local optimum is efficiently
computable by the local search method, it is also computable in
expected polynomial time by a GA with tournament selection.

In the present paper, the genetic algorithms are studied on a wide
class of combinatorial optimization problems. The expected number
of tentative solutions constructed, until first visiting a desired
area for the first time, is considered as the main criterion of GA
efficiency. Such an area may consist of locally optimal solutions
or of globally optimal solutions or of feasible solutions with
sufficiently small relative error. The main result is obtained by
combining the approaches from~\cite{Er12,L11} and applies to a
wider range of selection operators, compared to the result
from~\cite{Er12}, including the proportional selection of
Canonical GA~\cite{Gold} (the term ``Canonical GA'' was coined
in~\cite{Rud}). Considering the selection operators with very high
selection pressure, in this paper, we can neglect the probability
of downgrading mutations although this probability needs to be
taken into account in~\cite{CDEL14,DL16}. By downgrading mutations
here we mean mutations which decrease the quality of solutions (a
formal definition of downgrading mutation will be given further).
In contrast to~\cite{CDEL14,DL16}, here we consider explicitly the
constrained optimization problems and the expected first hitting
time of the set of local optima.

In most general setting, a combinatorial optimi\-za\-tion problem
with maximization criterion is formulated as follows:
\begin{equation} \label{eqn:problem}
\max\{F(x) \ | \ x \in {\rm Sol}\},
\end{equation}
where ${\rm Sol} \subseteq {\mathcal X}$ is the set of feasible
solutions, ${\mathcal X}=\{0,1\}^n$ is the search space,
$F(\cdot)$ is the objective function. The optimal value of the
criterion is denoted by~$F^*$. The minimization problems are
formulated analogously. Without loss of generality, by default we
will consider the maximization problems. The results will hold for
the minimization problems as well.

\paragraph{Genetic Algorithms.} In the process of the GA execution,
a sequence of populations~$P^t=({x}^{1,t},\dots,{x}^{\lambda,
t})$, \ $t=0,1,\dots,$ is computed, where each population consists
of~$\lambda$ {\it genotypes}. In the present paper, by the
genotypes we mean the elements of the search space~$\mathcal X$,
and {\em genes}~$x_i, \ i\in [n]$ are the components of a
genotype~$x \in \mathcal X$. Here and below, we use the
notation~$[n] := \{1, 2,...,n\}$ for any positive integer~$n$.

An initial population~$P^{0}$ consists of randomly generated
genotypes, and every next population is constructed on the basis
of the previous one. For convenience of GA description, in what
follows we assume that the population size~$\lambda$ is even.

In each iteration of a GA, $\lambda/2$ pairs of parent genotypes
are chosen from the current population~$P^t$ using the randomized
{\em selection} procedure~${\mbox{Sel}: {\mathcal X}^{\lambda} \to
[\lambda]}$. In this procedure, a parent genotype is independently
drawn from the previous population~$P^t$ where each individual is
assigned a selection probability depending on its {\em
fitness}~$f({x})$. Usually a higher fitness value of an individual
implies higher (or equal) selection probability. We assume that
the fitness function is defined on the basis of objective
function. If $\x\in {\rm Sol}$ then $f({x}) = \phi(F({x})),$ where
$\phi:\mathbb{R} \to \mathbb{R}_+$ is a monotone increasing
function in the case of maximization problem or a monotone
decreasing function in the case of minimization problem. Otherwise
(i.e. if ${x}\not \in {\rm Sol}$), the fitness incorporates some
penalty, which ensures that $f({x}) < \min_{\y \in {\rm Sol}}
F(\y).$

Given the current population, each pair of offspring genotypes is
created independently from other pairs using the randomized
operators of {\em crossover} and {\em mutation}. Some authors
consider crossover operators that output a single genotype (see
e.g.~\cite{CDEL14, BCh96, OW14, KP12,Vose}), while others consider
crossovers with two output genotypes (see e.g.~\cite{BN98, Gold,
Rud, V00}). For the sake of uniform treatment of both versions of
crossover, let us denote the number of output genotypes by~$r, \
r\in\{1,2\}$. In what follows, we assume that ${\mbox{Cross}:
{\mathcal X} \times {\mathcal X} \to {\mathcal X}^r}$ and
${\mbox{Mut}:{\mathcal X} \to {\mathcal X}}$ are efficiently
computable by randomized routines. When a new population~$P^{t+1}$
is constructed, the non-elitist GA proceeds to the next
iteration.\\
%

\noindent {\bf Algorithm 1. Non-Elitist Genetic Algorithm in the case of $r=2$}\\

\noindent Generate the initial population~$P^0$, assign $t:=1.$\\
{\bf While} a termination condition is not met {\bf do:}\\
$\mbox{\hspace{2em}}$ {\bf Iteration $t$.}\\
$\mbox{\hspace{2em}}$ {\bf For $j$ from 1 to ${\lambda}/2$ do:} \\
$\mbox{\hspace{4em}}$   Selection: $i:={\rm Sel}(P^{t})$, $i':={\rm Sel}(P^{t})$. \\
$\mbox{\hspace{4em}}$   Crossover: $(x,y):=\mbox{Cross}(x^{it},x^{i't}).$\\
$\mbox{\hspace{4em}}$   Mutation: $x^{2j-1,t+1} :=
\mbox{Mut}(x),\ \ x^{2j,t+1}:=\mbox{Mut}(y).$\\
$\mbox{\hspace{2em}}$ {\bf End for.}\\
$\mbox{\hspace{2em}}$ $t:=t+1.$\\
{\bf End while.}\\


\noindent {\bf Algorithm 2. Non-Elitist Genetic Algorithm in the case of $r=1$}\\

\noindent Generate the initial population~$P^0$, assign $t:=1.$\\
{\bf While} a termination condition is not met {\bf do:}\\
$\mbox{\hspace{2em}}$ {\bf Iteration $t$.}\\
$\mbox{\hspace{2em}}$ {\bf For $j$ from 1 to ${\lambda}$ do:} \\
$\mbox{\hspace{4em}}$   Selection: $i:={\rm Sel}(P^{t})$, $i':={\rm Sel}(P^{t})$. \\
$\mbox{\hspace{4em}}$   Crossover: $x:=\mbox{Cross}(x^{it},x^{i't}).$\\
$\mbox{\hspace{4em}}$   Mutation: $x^{j,t+1} := \mbox{Mut}(x).$\\
$\mbox{\hspace{2em}}$ {\bf End for.}\\
$\mbox{\hspace{2em}}$ $t:=t+1.$\\
{\bf End while.}\\

The output of a GA is an individual with the maximum fitness value
in all populations constructed until the termination condition was
met.

In the theoretical analysis of GAs it is often assumed that the
algorithm constructs an infinite sequence of populations and the
termination condition is never met. In practice, the termination
condition is required not only to stop the search and output the
result, but also to perform multiple restarts of the GA with
random initialization~\cite{BN98,BEGKK15}. Multiple independent
runs of randomized algorithms or local search (multistart) are
widely used to prevent localization of the search in the
``unpromising'' areas of the search space (see e.g.~\cite{BHM})
and applicability of multistart to the evolutionary algorithms has
some theoretical basis~\cite{DL15,V00}.

In this paper, together with the standard version of Non-Elitist
GA (Algorithms~1 and~2), we study the {\em GA with multistart},
where a GA outlined as Algorithm~1 or~2 is ran independently from
the previous executions for an unlimited number of times. The
stopping criterion in Algorithm~1 or~2 in this case is the
iterations limit $t\le t_{\max}$, where~$t_{\max}$ is a tunable
parameter.

In what follows, we consider three options for selection operator:
the tournament selection~\cite{Gold90}, the
$(\mu,\lambda)$-selection~\cite{L11} and the proportional
selection~\cite{Gold}. In \emph{$k$-tournament selection}, $k$
individuals are sampled uniformly at random with replacement from
the population, and the fittest of these individuals is returned.
The tunable parameter~$k$ is called the {\em tournament size}.
In $(\mu,\lambda)$-\emph{selection}, parents are sampled uniformly
at random among the fittest $\mu$ individuals in the
population~$P^t$. In the case of proportional selection,
\begin{equation}\label{eq:Psel}
\Pr({\rm Sel}(P^t)=i):=\frac{f(x^{it})}{\sum_{j=1}^{\lambda}
f(x^{jt})},
\end{equation}
if $\sum_{j=1}^{\lambda} f(x^{jt})>0$; otherwise the index of the
parent individual is chosen uniformly at random.

{\em Canonical Genetic Algorithm} proposed in~\cite{Gold}
corresponds to the GA outline with $r=2$, where all individuals of
the initial population are chosen independently and uniformly
from~${\mathcal X}$. This GA uses the proportional selection, a
{\em single-point crossover} $\mbox{\rm Cross}^*$ and  a {\em
bitwise mutation} $\mbox{\rm Mut}^*$. The last two operators work
as follows.

The single-point crossover computes
$({x}',{y}')=\mbox{Cross}^*({x},{y})$ for two input genotypes
${x=(x_1,...,x_n),}$ ${y=(y_1,..., y_n),}$ so that with a given
probability~$p_{\rm c}$,
$$
{x}'=(x_1,...,x_{\chi}, y_{\chi+1},...,y_n ), \ \ {y}'=(y_1,...,
y_{\chi}, x_{\chi+1},..., x_n),
$$
where the random position~$\chi$ is chosen uniformly from~1
to~$n-1$. With probability ${1-p_{\rm c}}$ both parent individuals
are copied without any changes, i.e. ${x'=x}, \ {y'=y}$.

The bitwise mutation $\mbox{\rm Mut}^*$ computes a
genotype~$x'=\mbox{\rm Mut}^*(x)$, where in\-dependent\-ly of
other bits, each bit~$x'_i, \ i\in [n]$, is assigned a
value~$1-x_i$ with probability~$p_{\rm m}$ and with
probability~$1-p_{\rm m}$ it keeps the value~$x_i$. The tunable
parameter~$p_{\rm m}$ is also called the {\em mutation rate}.
Choosing the mutation rate, many authors assume $p_{\rm m}=1/n.$

Another well-known operator of {\em point mutation} with a given
probability~$p_{\rm m}$ modifies one randomly chosen bit,
otherwise (with probability~$1-p_{\rm m}$) the given genotype
remains unchanged.

The following condition holds for many well-known crossover
operators: there exists a positive constant~$\varepsilon_0$ which
does not depend on the problem instance, such that the output of
crossover $({x}',{y}')=\mbox{Cross}({x},{y})$ satisfies the
inequality
\begin{equation}\label{eps_cross}
\varepsilon_0 \le \Pr\Big(\max\{f({x}'),f({y}')\} \ge
\max\{f({x}),f({y})\}\Big).
\end{equation}
for any ${x},{y} \in {\mathcal X}$. Condition~(\ref{eps_cross})
suggests that the fitness of at least one of the genotypes
resulting from crossover $({x}',{y}')=\mbox{Cross}({x},{y})$ is
not less than the fitness of the parents ${x},{y} \in {\mathcal
X}$ with probability at least~$\varepsilon_0$. This condition is
fulfilled for the single-point crossover with $\varepsilon_0 =
1-p_{\rm c}$, if $p_{\rm c}<1$ is a constant. In the case of
crossover operator with a single output genotype
${x}'=\mbox{Cross}({x},{y})$ analogous condition is as follows
\begin{equation}\label{eps_cross1}
\varepsilon_0 \le \Pr\Big(f({x}') \ge \max\{f({x}),f({y})\}\Big).
\end{equation}
Condition~(\ref{eps_cross1}) is also satisfied with $\varepsilon_0
= 1$ for the optimized crossover operators, where at least one of
the two offspring is computed as a solution to optimal
recombination problem (see e.g.,~\cite{AOT97,BN98,EK14}). It was
shown in~\cite{CDEL14} that for some well-known crossover
operators and simple fitness functions
condition~(\ref{eps_cross1}) holds with $\varepsilon_0 = 1/2$.

\section{THE MAIN RESULT}

This section generalizes the analysis of Non-Elitist Genetic
Algorithm carried out in~\cite{Er12}, adapting it to different
selection operators and making it applicable to the GAs with
multistart, which allows us to deal with both feasible and
infeasible solutions.

Suppose that for some~$m$ there is an ordered partition
of~$\mathcal{X}$ into subsets $A_0,\dots,A_{m+1}$ called {\em
levels}~\cite{CDEL14}. Level~$A_{0}$ may be an empty set.
Level~$A_{m+1}$ will be the target level in subsequent analysis.
The target level may be chosen as the set of solutions with
maximal fitness or the set of local optima or the set of
$\rho$-approximation solutions for some approximation factor~$\rho
> 1$ (a feasible solution~$y$ to a maximization problem is
called a $\rho$-approximation solution if it satisfies the
inequality $F^*/F({y}) \leq \rho$). A well-known example of
partition is the \emph{canonical} partition, where $A_0=\emptyset$
and each level~$A_j, \ j\in[m+1]$ regroups solutions having the
same fitness value (see e.g.~\cite{DL16,S10}). In what follows,
level~$A_0$ may be used to encompass the set of infeasible
solutions.

In this paper, we will often use values which are independent of
an instance of problem~(\ref{eqn:problem}) and of a levels
partition, but completely determined by the GA outline and its
operators. Such values will be called constants. The same applies
to the constants in~$O(\cdot)$ notation. It will be convenient to
use the symbol $H_j:=\cup_{i=j}^{m+1} A_i$ for the union of all
levels starting from level~$j,$ $j\in [m+1]$. The symbol~$e$ in
what follows denotes the base of the natural logarithm.


Extending the notation from~\cite{CDEL14,L11}, we will define the
\emph{selective pressure}~$\beta(0,P)$ of a selection
operator~${\rm Sel}(P)$ as the probability of selecting an
individual that belongs to the highest level occupied by the
individuals of~$P$.

\begin{theorem}\label{theor:GA_probab} Given a partition
  $A_0,\ldots,A_{m+1}$ of~$\mathcal{X}$, let there
  exist parameters $s_*,p_1,\varepsilon$ and~$\beta_0$ from $(0,1]$,
  such that for any $j\in[m]$:
  \begin{description}
  \item[(C1)] ${\Pr}(\mbox{Mut}({x})\in H_{j+1})\geq s_*$ for any $x\in A_{j},$
  \item[(C2)] $\Pr\left(x^{i,0} \in H_1 \ \mbox{for some}\ i\in [\lambda] \right) \ge p_1,$
  \item[(C3)] $\beta(0,P)\geq \beta_0$
  for any $P\in (\mathcal{X}\backslash A_{m+1})^\lambda$,
  \item[(C4)] $\lambda \geq
    \frac{2(1+\ln m)}{s_* \varepsilon \beta_0(2-\beta_0)}$,
  \item[(C5)] for any
           $(x,y) \in (H_j \times \mathcal{X}) \cup (\mathcal{X} \times H_j)$
$$
 \varepsilon \le
 \left\{
\begin{array}{ll}
\Pr\Big(\mbox{Cross}(x,y) \in H_j\Big), & \mbox{\rm in case of } \ \ r=1,\\
\Pr\Big(\mbox{Cross}(x,y) \in
          (H_j \times \mathcal{X}) \cup (\mathcal{X} \times H_j)\Big), &
          \mbox{\rm in case of } \ \ r=2.
\end{array} \right.
$$

  \end{description}
  Then with probability not less than~$p_1/e$ at least one
  of the populations~$P^0,P^1,\dots,P^m$ contains an individual
  from~$A_{m+1}$.
\end{theorem}

Let us informally describe the conditions of the theorem.
Condition~(C1) requires that for each level~$j$, there is a lower
bound~$s_*$ on the ``upgrade'' probability \ from level~$j$.
Condition~(C2) ensures that at least one individual of the initial
population is above \ level~0 with probability not less
than~$p_1$. Condition~(C3) requires that the selective pressure
induced by the selection mechanism is sufficiently high.
Condition~(C4) requires that the population size~$\lambda$ is
sufficiently large. Condition (C5) is a level-based analog of
inequalities~(\ref{eps_cross}) and~(\ref{eps_cross1}). This
condition follows from~(\ref{eps_cross}) or~(\ref{eps_cross1})
with $\varepsilon=\varepsilon_0$ in the case of the canonical
partition.


{\bf Proof of Theorem~\ref{theor:GA_probab}.} For any
$t=0,1,\dots$ let the event~$E_i^{t+1}$, $i\in [\lambda/2]$,
consist in fulfilment of the following three conditions when the
$i$-th pair of offspring is computed:
\begin{enumerate}
\item At least one
of the two parents is chosen from the highest level~$A_j$ occupied
by population~$P^t$.
 \item When the crossover operator is
applied, at least one of its outputs belongs to~$H_j$. W.l.o.g. we
assume that this output is~$x$.
 \item The mutation operator applied to~${x}$
produces a genotype in~$H_{j+1}$.
\end{enumerate}

Let~$p$ denote the probability of the union of events~$E_i^{t+1},
\ i\in [\lambda/2]$. In what follows, we will construct a lower
bound~$\ell \le p$, which holds for any population~$P^t$.
According to the outline of GA, $\Pr(E_1^{t+1})=\dots =
\Pr(E_{\lambda/2}^{t+1})$. Let us denote this probability by~$q$.
Note that~$q$ is bounded from below by $s_* \varepsilon
(1-(1-\beta_0)^2)=s_* \varepsilon\beta_0(2-\beta_0).$ Given a
population~$P^t$, the events $E_j^{t+1}, \ j=1,\dots,\lambda/2,$
are independent, so $p \ge 1-(1-q)^{\lambda/2} \ge 1-e^{-q
\lambda/2}$. In what follows we shall use the fact that
condition~(C4) implies
\begin{equation} \label{useful}
\lambda \ge \frac{2}{s_* \varepsilon \beta_0(2-\beta_0)} \ge 2/q.
\end{equation}
To bound probability~$p$ from below, we first note that for any $z
\in [0,1]$ holds
\begin{equation}\label{simple} 1-\frac{z}{e} \ge e^{-z}.
\end{equation}
Assume $z=e^{-q \lambda/2+1}$. Then in view of
inequality~(\ref{useful}), $z\le 1$, and consequently,
\begin{equation}\label{lower_bound_on_p}
p \ge \exp\left\{-e^{1-q \lambda/2}\right\} \ge
\exp\left\{-e^{1-s_* \varepsilon
\beta_0(2-\beta_0)\lambda/2}\right\}.
\end{equation}
We will use the right-hand side expression
of~(\ref{lower_bound_on_p}) as the lower bound~$\ell$ for~$p$.

For any $t=1,2,\dots$ let us define the
event~$\mathcal{E}_{t}:=E_1^{t}+\dots+E_{\lambda/2}^{t}$. Note
that event~$\mathcal{E}_{t}$ captures some of the possible
scenarios of ``upgrading'' the best individuals of the current
population to the next level. Besides that, let $\mathcal{E}_{0}$
denote the event that $x^{i,0} \in H_1$ for some $i\in [\lambda]$.
Then the probability to reach the target level~$A_{m+1}$ in a
series of at most~$m$ iterations is lower bounded by
${\Pr}(\mathcal{E}_0\& \dots \& \mathcal{E}_m)$ and
\begin{equation}\label{cond_probab}
{\Pr}(\mathcal{E}_0\& \dots \& \mathcal{E}_m) =
{\Pr}(\mathcal{E}_0) \prod_{t=0}^{m-1}
{\Pr}(\mathcal{E}_{t+1}|\mathcal{E}_0 \& \dots \& \mathcal{E}_{t})
\ge p_1 \ell^m.
\end{equation}
in view of condition~(C2). Now using condition~(C4), we get:
$$
\ell^m = \exp\left\{-m e^{1-s_* \varepsilon \beta_0(2-\beta_0)
\lambda/2}\right\} \ge
 \exp\left\{-m e^{-\ln m}\right\}=1/e.
$$
$\Mybox$\\

An event of {\em downgrading mutation} of a genotype~$x$ may be
defined in terms of levels partition as $\mbox{Mut}({x})\not\in
H_{j}$, where $j$ is the level the individual~$x$ belongs to.
Unlike the results from~\cite{CDEL14,DL16},
Theorem~\ref{theor:GA_probab} is applicable to the GAs where the
probability of non-downgrading mutation may tend to zero as the
problem size grows. Examples of such operators may be found in
highly competitive GAs for Maximum Independent Set Problem and Set
Covering Problem~\cite{AOT97,BCh96} and many other GAs in the
literature on operations research. Note that in case
$|A_{m+1}|=1$, given an optimal genotype $x\in A_{m+1}$, the
bitwise mutation with a constant mutation rate (as used
in~\cite{AOT97}) causes non-downgrading mutations only with
probability~$(1-p_{\rm m})^n=o(1)$ and the mutation operator that
inverts~$m_f$ bits, where~$m_f>0$ is a given
parameter~\cite{BCh96}, has zero probability of non-downgrading
mutations.

\section{LOWER BOUNDS FOR SELECTION PRESSURE}

The following two propositions may be applied to check
condition~(C3) in Theorem~\ref{theor:GA_probab}.

\begin{proposition} \label{prop:tournament-selection}
  Let levels $A_1,\dots,A_m$ satisfy the monotonicity condition
\begin{equation}\label{eqn:relaxed_monoton}
f(x)<f(y) \ \mbox{for any} \ x\in A_{j-1}, \ y \in A_j, \
j=2,\dots,m.
\end{equation}
  Then

  (i) $k$-tournament selection
  with $k \ge \alpha \lambda$, where the constant $\alpha>0$,
  satisfies condition~(C3) with ${\beta_0=1-e^{-\alpha}}$.

  (ii) $(\mu,\lambda)$-selection with a constant parameter~$\mu\le \lambda$
  satisfies condition~(C3) where ${\beta_0=1/\mu}$.
\end{proposition}

{\bf Proof.} In the case of $k$-tournament selection~$ \beta(0,P)
\ge 1-\left(1-1/\lambda\right)^{k}$ and $(1-1/\lambda)^{k} \le
(1-1/\lambda)^{\alpha\lambda} \le e^{-\alpha}$, so part~(i)
follows. Part~(ii) follows from the definition of
$(\mu,\lambda)$-selection immediately. $\Mybox$\\

The operator of proportionate selection does not have a tunable
parameter that allows to set its selection pressure. However such
a parameter (let it be~$\nu$) may be introduced into the fitness
function by assuming that~$f(x)=F(x)^{\nu}$ for any~$x\in{\rm
Sol}$. The proof of the following proposition is similar to that
of Lemma~8 in~\cite{L11}. Here and below $\Z_+$ denotes the set of
non-negative integers.

\begin{proposition}\label{prop:prop}
Let the levels $A_1,\dots,A_m$ satisfy the monotonicity
condition~(\ref{eqn:relaxed_monoton}), $F: {\rm Sol} \to
\mathbb{Z}_+$ and the fitness function is of the
form~$f(x)=F(x)^{\nu}$, where $\nu>\max(0,\ln(\alpha \lambda)F^*)$
for some $\alpha>0$. Then the proportional selection
  satisfies condition~(C3) with $\beta_0=1/(1+\alpha^{-1})$.
\end{proposition}

\noindent {\bf Proof.} Let~$F_0^{\nu}$ be the maximal fitness
value in population~$P$ and let $k$ denote the number of
individuals in~$P$ with fitness~$F_{0}^{\nu}$. The probability to
choose one of the fittest individuals is lower bounded as follows
$$
\beta(0,P) \ge \frac{kF_0^{\nu}}{(\lambda-k)(F_{0}-1)^{\nu}
+kF_0^{\nu}}
 \ge \frac{k}{\lambda(1-1/F_{0})^{\nu} +k}
 \ge \frac{k}{1/\alpha+k}  \ge \frac{1}{1/\alpha+1},
$$
since $(1-1/F_{0})^{\nu}\le (1-1/F^*)^{\nu}\le e^{-\nu/F^*} \le
1/(\alpha \lambda).$ $\Mybox$\\

Proposition~\ref{prop:prop} requires the fitness function to scale
very fast as the objective function grows. Scaling of objective
function might be unavoidable in the case of proportional
selection. Even for the simple benchmark fitness function
$\OM:=\sum_{i=1}^n x_i$, P.~Oliveto and C.~Witt show~\cite{OW14}
that in the case of proportional selection, GA with high
probability makes exponential number of iterations until the
optimum is visited. The need for scaling the fitness function is
also acknowledged in practical use of Canonical GA (see
e.g.~\cite{Gold}, where a dynamical mechanism for fitness scaling
was proposed).

\section{UPPER BOUNDS ON EXPECTED HITTING TIME OF TARGET SUBSET}

Let $T$ denote the random variable, equal to the number of
tentative solutions evaluated until some element of the current
population is sampled from~$A_{m+1}$ for the first time. In the
case when $A_{m+1}$ is the set of optimal solutions, $T$ is
usually called the {\em runtime} of an evolutionary algorithm.

\begin{corollary} \label{cor:Sol0}
Suppose that conditions~(C1)-(C5) of Theorem~\ref{theor:GA_probab}
hold and $A_{0}=\emptyset$. Then for the GA we have $E[T] \le
em\lambda.$
\end{corollary}

{\bf Proof.} Consider a sequence of series of the GA iterations,
where the length of each series is~$m$ iterations. Suppose, $D_i,
\ i=1,2,\dots,$ denotes an event of absence of solutions
from~$A_{m+1}$ in the population throughout the $i$-th series. The
probability of each event~$D_i, \ i=1,2,\dots,$ is at most~$1-1/e$
according to Theorem~\ref{theor:GA_probab}. Analogously to
bound~(\ref{cond_probab}) we obtain the inequality ${\Pr}(D_1\&
\dots \& D_i) \le (1-1/e)^i.$

Let~$Y$ denote the random variable, equal to the number of the
first run when a solution form~$A_{m+1}$ was obtained. By the
properties of expectation (see e.g.~\cite{Gnedenko}),
$$
E[Y] = \sum_{i=0}^{\infty} {\Pr}(Y > i) = 1+\sum_{i=1}^{\infty}
{\Pr}(D_1\& \dots \& D_i) \le 1 +\sum_{i=1}^{\infty} (1-1/e)^i =
e.
$$
Consequently, the average number of iterations until an element of
the target subset is first obtained is at most~$em$.
$\Mybox$\\

Assuming $\lambda =\left\lceil
    \frac{2(1+\ln(m))}{s_* \varepsilon
    \beta_0(2-\beta_0)}\right\rceil$ and constant
    $\beta_0$ and $\varepsilon$, Corollary~\ref{cor:Sol0} implies $E[T]
\le c m\ln(m)/s_*$, where $c > 0$ is a constant. In the special
case where $r=1$ and the probability of non-downgrading mutation
${\Pr}(\mbox{Mut}({x})\in H_{j}\ | \ x\in A_j), \ j\in[m]$ is
lower bounded by a positive constant, the result
from~\cite{CDEL14} gives an upper bound $\E[T] \leq
c'm\left(\ln(m/s_*)\ln\ln(m/s_*) +{1}/s_*\right)$ with some
positive constant~$c'$. The latter bound is less demanding to
selection pressure and it is asymptotically tighter than the bound
$E[T] \le c m\ln(m)/s_*$ e.g. when $s_*\le 1/m$.

Note that the assumption $A_0=\emptyset$ in
Corollary~\ref{cor:Sol0} can not be dismissed. Indeed, suppose
that $A_0\not=\emptyset$,
and consider a~GA where the mutation operator has the following
properties. On one hand, it never outputs an offspring in~$H_1$,
given an input from~$A_0$. On the other hand, given a
genotype~$x\in H_1$, the result of mutation is in~$A_0$ with a
probability at least $c$, where $c>0$. Finally assume that the
initialization procedure produces no genotypes from~$A_{m+1}$ in
population~$P^0$ and the crossover makes no changes to the parent
genotypes. Now all conditions of Corollary~\ref{cor:Sol0} can be
satisfied but with a positive probability of at
least~$c^{\lambda}$ the whole population~$P^1$ consists of
solutions from~$A_0$, and subject to this event all
populations~$P^1,P^2,\dots$ contain no solutions from~$H_1$.
Therefore, $E[T]$ is unbounded.

As an example of usage of Corollary~\ref{cor:Sol0} we consider the
GA with tournament selection applied to the family of
unconstrained optimization problems with objective
function~$\leadingones$, which is frequently used in the analysis
of evolutionary algorithms. The objective function $\leadingones:
\{0,1\}^n \to \Z_+$ is defined as
$$
\leadingones (\x)= \sum_{i=1}^{n} \prod_{j=1}^{i} x_j
$$
i.e. the optimal solution is $x^*=(1,\dots,1)$.

Let us use the canonical levels partition: $A_j=\{x\ | \
F(x)=j-1\}$, ${j \in [n+1]}$, $m=n$. Assume that the bitwise
mutation operator has the mutation rate~$p_{\rm m}=1/n$. To move
from level~$A_{j}$ to level~$A_{j+1}$ under mutation it suffices
to modify the first zero bit and not to modify the rest of the
bits. So we can use $s_*={(1/n)(1-1/n)^{n-1}}=\Omega(1/n)$.
Suppose that in the single-point crossover~$p_{\rm c}=1$. Then in
the case of~\leadingones, as it was shown in~\cite{CDEL14}, the
constant $\epsilon=1/2$ satisfies condition~(C5). Assuming the
tournament size $k = \Theta(\lambda)$,
Proposition~\ref{prop:tournament-selection} ensures satisfaction
of condition~(C3) with a positive constant~$\beta_0$.

Application of Corollary~\ref{cor:Sol0} to the GA with~$r=2$ and
$\lambda=\Theta(n\ln(n)),$ satisfying~(C4), gives the upper bound
$E[T]=O(n^2 \ln(n))$. Note that a similar bound obtained
in~\cite{CDEL14} for the case of single-offspring crossover
gives~$E[T]=O(n^2).$ We expect that extension of the drift
analysis~\cite{CDEL14} to the case of two-offspring crossovers may
tighten the upper bound for the \leadingones in the case of~$r=2$
as well.

Analogously to Corollary~\ref{cor:Sol0} we obtain

\begin{corollary} \label{cor:Sol}
Let the GA with multistart use the termination condition
with~$t_{\max}=m$. Then $E[T] \le em\lambda/p_1$ holds under
conditions~(C1)-(C5).
\end{corollary}

As an illustrative example for Corollary~\ref{cor:Sol} we consider
Canonical GA on the family of instances of Set Cover Problem
proposed by E.Balas in~\cite{Ba84}. In general the set cover
problem~(SCP) is formulated as follows. Given: $M=\{1,...,m\}$ and
a set of subsets $M_j\subseteq M$, $j\in [n]$. A subset $J
\subseteq [n]$ is called a {\it cover} if $\cup_{j \in J}
{M_j}=M.$ The goal is to find a cover of minimum cardinality. In
what follows we denote by~$N_i$ the set of indices of the subsets
that cover an element~$i$, i. e. $N_i=\{j: i \in M_j\}$ for any
$i$.

In the family ${\mathcal B}(n,p)$ of SCPs introduced by E.~Balas
in~\cite{Ba84}, it is assumed that $m=C_{n}^{{p}-1}$ and the set
$\{N_1,N_2,...,N_{m}\}$ consists of all $(n-p+1)$-element subsets
of $[n]$. Thus $J \subseteq [n]$ is an optimal cover iff $|J|=p$.

Family ${\mathcal B}(n,p)$ is known to have a large fractional
cover~\cite{Ba84} which implies that these SCPs are likely to be
hard for integer programming methods. In particular, it was shown
in~\cite{Za} that problems from this class are hard to solve using
the $L$-class enumeration method~\cite{Kolo}. When~$n$ is even
and~$p=n/2$, the $L$-class enumeration method needs an exponential
number of iterations in~$n$. In what follows, we analyze GA in
this special case.

In the binary encoding of solutions we assume that each bit
$x_j\in\{0,1\}, j \in [n],$ indicates whether~$j$ belongs to the
encoded set, $J(x):=\{j\in [n] : x_j=1\}$. If $J(x)$ is a cover
then we assume $F(x)={n}-|J(x)|+1$, otherwise we put $F(x)=0$ as a
penalty.

Consider Canonical GA with multistart and scaled fitness
function~$f(x)=F(x)^{\nu}$, the termination condition
where~$t_{\max}=n/2$, a constant parameter~$p_{\rm c}<1$ and the
mutation rate~$p_{\rm m}=1/n$.

Assume that~$A_0$ is the set of all infeasible solutions and the
rest of the levels $A_1,\dots,A_{m+1}$ are defined according to
the canonical partition on~${\rm Sol}$, where $m=n/2$.
In the case of $p=n/2$, with probability~1/2 a random individual
of~$P^0$ is feasible and there exists a constant~${p_1>0}$
satisfying condition~(C2). The constant $\epsilon=1-p_{\rm c}$
satisfies condition~(C5). The probability that under mutation a
genotype from level~$A_j$ produces an element of~$H_{j+1}, \
j\in[m]$ in the case of problems of family~${\mathcal B}({n},{p})$
is lower bounded by~$s_*=\Omega(1)$. Choosing $\nu>\ln(\alpha
\lambda)n/2$ with constant $\alpha>0$ we ensure condition~(C3)
according to Proposition~\ref{prop:prop}. Finally, appropriate
$\lambda=\Theta(\ln(n)),$ satisfies condition~(C4). Therefore
Corollary~\ref{cor:Sol} implies that an optimal solution is
attained for the first time after~$\E[T]=O\left(n\ln(n)\right)$
tentative solutions in expectation.

\section{APPLICATIONS TO LOCAL SEARCH PROBLEMS}
\label{sec:unconstrained}

In this section, GAs are compared to the local search method. In
order to keep track of running times w.r.t. the length of problem
instance encoding, here the combinatorial optimization problems
are viewed under the technical assumptions of the class of
NP~optimization problems (see e.g.~\cite{ACGKMP}). Let $\{0,1\}^*$
denote the set of all strings with symbols from~$\{0,1\}$ and
arbitrary string length. For a string~$S\in \{0,1\}^*$, the
symbol~$|S|$ will denote its length. In what follows, $\N$ denotes
the set of positive integers and given a string~$S\in \{0,1\}^*$,
the symbol~$|S|$ denotes the length of the string~$S$. To denote
the set of polynomially bounded functions we define $\mbox{\rm
Poly}$ as the class of functions from $\{0,1\}^*$ to $\N$ bounded
above by a polynomial in~$|I|,$ where $I \in \{0,1\}^*$.

\begin{definition}\label{def:NPO} An NP~optimization problem $\Pi$
is a triple ${\Pi=(\mbox{\rm Inst},\mbox{\rm Sol}(I),F_I)}$, where
$\mbox{\rm Inst} \subseteq \{0,1\}^*$ is the set of instances
of~$\Pi$ and:

1. The relation $I \in \mbox{\rm Inst}$ is computable in
polynomial time.

2. Given an instance $I \in \mbox{\rm Inst}$, $\mbox{\rm
Sol}(I)\subseteq \{0,1\}^{n(I)}$ is the set of feasible solutions
of~$I$, where~$n(I)$ stands for the dimension of the search
space~${\mathcal X}_I:=\{0,1\}^{n(I)}$. Given $I \in \mbox{\rm
Inst}$ and $\x \in \{0,1\}^{n(I)}$, the decision whether $\x\in
\mbox{\rm Sol}(I)$ may be done in polynomial time, and $n(\cdot)
\in \mbox{\rm Poly}$.

3. Given an instance~$I \in \mbox{\rm Inst}$,  $F_I: \mbox{\rm
Sol}(I) \to {\N}$ is the objective function (computable in
polynomial time) to be maximized if $\ \Pi$ is an NP~maximization
problem or to be minimized if $\ \Pi$ is an NP~minimization
problem.
\end{definition}

The symbol of problem instance~$I$ may often be skipped in the
notation, when it is clear what instance~$I$ is meant. A
combinatorial optimization problem $\Pi=(\mbox{\rm Inst},\mbox{\rm
Sol}(I),F_I)$ is called polynomially bounded, if there exists a
polynomial in~$|I|$, which bounds the objective values $F_I({x})$,
${x \in {\rm Sol}(I)}$ from above.

Let a neighborhood ${\mathcal N}({y})\subseteq \mbox{\rm Sol}$ be
defined for every~${y}\in \mbox{\rm Sol}$. The mapping ${\mathcal
N}: {\rm Sol} \to 2^{{\rm Sol}}$ is called the {\em neighborhood
mapping}. The family~$\{{\mathcal N}({y}): {y} \in \mbox{\rm
Sol}\}$ is called the {\em neighborhoods system}. One of the
standard neighborhoods systems on~$\mbox{\rm Sol}=\{0,1\}^n$ is
{\em Hamming neighborhoods system}: ${\mathcal N}({y})=\{x \ | \
d(x,y)\le \ R\}, \ y\in \mbox{\rm Sol},$ where the radius~$R$ is a
constant and $d(\cdot,\cdot)$ denotes the Hamming distance. If the
inequality $F({y})\leq F({x})$ holds for all neighbors~$\y \in
{\mathcal N}({x})$ of a solution~${{x} \in \mbox{\rm Sol}}$,
then~${x}$ is called a local optimum w.r.t.~${\mathcal N}$. In
what follows, the set of all local optima is denoted
by~$\mathcal{LO}$.

A local search method starts from some feasible solution~$\y_0$.
Each iteration of the algorithm consists in moving from the
current solution to a new solution in its neighborhood, such that
the value of objective function is increased. The way to choose an
improving neighbor, if there are several of them, will not matter
in this paper. The algorithm continues until it will reach a local
optimum.

Suppose that some neighborhood system~$\mathcal{N}$ is defined for
problem~(\ref{eqn:problem}) and $s$ is the lower bound for
probability that the mutation operator transforms a given
solution~$\x$ into a specific neighbor~${y} \in {\mathcal N}
({x})$, i.e.
\begin{equation}\label{eqn:s_def}
s \le {\Pr}(\mbox{Mut}({x})={y}), \ {x} \in {\rm Sol}, \ {y} \in
{\mathcal N}({x}).
\end{equation}

Many well-known combinatorial optimization problems, such as
Maximum Satisfiability Problem,  Maximum Cut Problem and Ising
Spin Glass Model~\cite{Barahon} have a set of feasible solutions
equal to the whole search space~${\rm Sol}\equiv \mathcal X$. The
following two corollaries apply to the problems with such a
property.

Let~$m$ be the number of different values of the fitness function
$f_1<\dots<f_{m}$ on~$\mathcal{X} \backslash \mathcal{LO}$, i.e.
$m=|\{g: g=f({x}), \ {x} \in \mathcal{X} \backslash \mathcal{LO}
\}|$. Then, starting from any initial solution, the local search
method attains a local optimum within at most~$m$ iterations. Let
us use a modification of the canonical levels partition, grouping
all local optima into the target subset~$A_{m+1}$:
\begin{equation} \label{eqn:partit_1}
A_j := \{x\in \mathcal{X} | f(x) = f_j\} \backslash \mathcal{LO},
\ j\in [m],
\end{equation}
\begin{equation} \label{eqn:partit_2}
A_{m+1} := \mathcal{LO}.
\end{equation}

Application of Corollary~\ref{cor:Sol0} and
Proposition~\ref{prop:tournament-selection} with levels
partition~(\ref{eqn:partit_1}), (\ref{eqn:partit_2}) gives

\begin{corollary}\label{cor:GA_LS1}
Suppose ${\rm Sol}\equiv{\mathcal X}$, a constant~$\epsilon_0>0$
satisfies inequality~(\ref{eps_cross}) or~(\ref{eps_cross1}),
$s>0$ satisfies inequality~(\ref{eqn:s_def}) and the GA uses a
$k$-tournament selection with $k>\alpha \lambda$ or
 $(\mu,\lambda)$-selection, where $\alpha$ and
  ${\mu}$ are constants.
  Then there exists a constant~$c>0$ such that a GA with population
  size~$\lambda \ge c\ln\left(m\right)/s$ first visits
  a local optimum of problem~(\ref{eqn:problem})
  after at most~$e\lambda m$ tentative
  solutions in expectation.
\end{corollary}

Therefore, with an appropriate population size, e.g. $\lambda
=\lceil c\ln(m)/{s}\rceil$, under conditions of
Corollary~\ref{cor:GA_LS1}, a local optimum of
problem~(\ref{eqn:problem}) is visited for the first time after
evaluation of at most~$e m \ln(m)/s$ tentative solutions on
average. This fact in the special case of the tournament selection
and $r=2$ was proved in~\cite{Er12}).

In order to consider bitwise mutation in more detail we will use
the following definition from~\cite{AP95}. A neighborhood
mapping~$\mathcal{N}$ is called {\em $K$-bounded}, if for any ${y}
\in {\rm \mbox{\rm Sol}}$ and ${x} \in \mathcal{N}({y})$ holds
$d({x},{y}) \le K$, where $K$ is a constant.

The bitwise mutation operator~$\mbox{Mut}^*$ outputs a
string~${x}$, given a string~${y}$, with probability $p_{\rm
m}^{d({x},{y})}(1-p_{\rm m})^{n-d({x},{y})}$. Note that
probability~$p_{\rm m}^j(1-p_{\rm m})^{n-j}$, as a function
of~$p_{\rm m}$, \ $p_{\rm m} \in [0,1]$, attains its minimum at
$p_{\rm m}=j/n$. The following proposition from~\cite{Er12} gives
a lower bound for the probability
$\Pr\{{\mbox{Mut}^*({y})={x}}\}$, which is valid for any ${x} \in
\mathcal{N}({y})$, assuming that~$p_{\rm m}=K/n$. We reproduce
this proposition here with a proof for the sake of completeness.

\begin{proposition}\label{optimal_bound}
Suppose the neighborhood mapping~$\mathcal{N}$ is $K$-bounded,
$K\le n/2$ and $p_{\rm m}=K/n$. Then for any ${y} \in {\rm
\mbox{\rm Sol}}$ and any ${x} \in \mathcal{N}({y})$ holds
${\Pr}\{\mbox{\rm Mut}^*({y})={x}\} \ge \left({K}/{en}\right)^K. $
\end{proposition}

{\bf Proof.} For any ${y} \in {\rm \mbox{\rm Sol}}$ and ${x} \in
\mathcal{N}({y})$ holds
$$
{\Pr}\{\mbox{Mut}^*({y})={x}\}
=\left(\frac{K}{n}\right)^{d({x},{y})}\left(1-\frac{K}{n}\right)^{n-d({x},{y})}
\ge
 \left(\frac{K}{n}\right)^K
 \left(1-\frac{K}{n}\right)^{n-K},
$$
since $p_{\rm m} =K/n\le 1/2$. Now $\frac{\partial}{
\partial n} (1-K/n)^{n-K} <0$ for $n>K$, and $(1-K/n)^{n-K} \to 1/e^K$
as~$n\to \infty$. Therefore $(1-K/n)^{n-K} \ge 1/e^K$. $\Mybox$\\

Application of Corollary~\ref{cor:Sol0} and
Propositions~\ref{prop:prop}~and~\ref{optimal_bound} to Canonical
GA yields

\begin{corollary}\label{cor:GA_LS2}
Suppose that ${\rm Sol}\equiv{\mathcal X}$, $F:{\rm Sol} \to
\mathbb{Z}_+$, a neighborhood mapping~$\mathcal{N}$ is
$K$-bounded, $p_{\rm c}<1$ is a constant, $p_{\rm m}=K/n$ and the
fitness function has a form~$f(x)=F(x)^{\nu}$, where
$\nu>\ln(\alpha \lambda)F^*$. Then there exists such
constant~$c>0$ that Canonical GA with~$\lambda \ge
c\ln\left(F^*\right)/n^K$ visits a local optimum to
problem~(\ref{eqn:problem}) for the first time after at
most~$eF^*\lambda$ tentative solutions in expectation.
\end{corollary}

Corollary~\ref{cor:GA_LS2} implies that in the case of
polynomially bounded unconstrained NP~optimization problem,
Canonical GA given appropriate choice of parameters finds a local
optimum in Hamming neighborhoods system within expected polynomial
time.

Let us consider a GA with multistart applied to an NP optimization
problem, i.e. in general ${\rm Sol}$ may be a proper subset
of~${\mathcal X}$. Corollary~\ref{cor:Sol} and
Proposition~\ref{prop:tournament-selection} yield

\begin{corollary} \label{cor:general}
Suppose that inequality~(\ref{eps_cross}) or~(\ref{eps_cross1})
holds for some constant $\epsilon_0>0$, bound~$s$ satisfies
inequality~(\ref{eqn:s_def}) and condition~(C2) is satisfied for
some constant~$p_1>0$. Besides that assume that GA with multistart
uses a termination condition $t_{\max}=m$ and one of the following
selection operators:
\begin{itemize}
\item $k$-tournament selection with
$k>\alpha\lambda$ where $\alpha>0$ is a constant or
\item $(\mu,\lambda)$-selection with a constant~$\mu$ or
\item proportional selection in the case of $F: {\rm Sol} \to
\mathbb{Z}_+$ and the fitness function is of the
form~$f(x)=F(x)^{\nu}$ where $\nu>\ln(\alpha\lambda)F^*$ and
$\alpha>0$ is a constant.
\end{itemize}
Then there exists such positive constant~$c$ that with population
size~$\lambda \ge c\ln\left({m}\right)/s$ a local optimum is first
reached by the GA with multistart after evaluation of at most~$e
m\lambda$ tentative solutions in expectation.
\end{corollary}

Corollary~\ref{cor:general} is formulated for the GA with
multistart rather than single-run GA because in general this
result does not hold for the single-run GA. Indeed, suppose ${\rm
Sol} \neq {\mathcal X}$ and consider a GA where the mutation
operator has the following properties. On one hand~${\rm Mut}$
never outputs a feasible offspring, given an infeasible input. On
the other hand, given a feasible genotype~$x$, ${\rm Mut}(x)$ is
infeasible with a {\em positive} probability, lower bounded by a
constant~$\epsilon\in(0,1]$. Finally assume that the
initialization procedure for population~$P^0$ produces only
feasible solutions, but none of them is locally optimal. Now all
conditions of Corollary~\ref{cor:general} are satisfied, but with
a positive probability of at least~$\epsilon^{\lambda}$ the whole
population~$P^1$ consists of infeasible solutions, and subject to
this event all populations~$P^1,P^2,\dots$ are infeasible.
Therefore if the GA is run without restarts, the expected number
of iterations until the first improvement of the best found
solution is unbounded and the expected hitting time of a local
optimum is unbounded as well. The need for restarting the GA was
overlooked in the first publication of a result analogous to
Corollary~\ref{cor:general} in~\cite{Er12}. The GA considered
in~\cite{Er12} should be replaced by the GA with multistart using
the termination condition $t_{\max}=m$ to make the results
in~\cite{Er12} correct in the case of ${\rm Sol} \neq {\mathcal
X}$. This correction is implemented in~\cite{Er13}.

Corollary~\ref{cor:general} may be used to estimate the capacities
of GAs to find efficiently the solutions with guaranteed
approximation ratio if all local optima of a problem have a known
approximation ratio.

\begin{definition}\label{GLO} {\rm \cite{AP95}}
A polynomially bounded NP~optimization problem~$\Pi$ belongs to
the class of {\em Guaranteed Local Optima}~(GLO) problems, if the
following two conditions hold:

1) At least one feasible solution~${y}_I \in {\rm \mbox{\rm Sol}}$
is efficiently computable for every instance~$I \in {\rm Inst}$;

2) A $K$-bounded neighborhood mapping~$\mathcal{N}_I$ exists, such
that for every instance~$I$, any local optimum of~$I$ with respect
to~$\mathcal{N}_I$ has a constant guaranteed approximation ratio.

\end{definition}

The class~{\rm GLO} contains such well-known {\rm NP}~optimization
problems as the Maximum Staisfiablity and the Maximum Cut
problems, besides that, on graphs with bounded vertex degree the
Independent Set problem, the Dominating Set problem and the Vertex
Cover problem also belong to GLO~\cite{AP95}.

If a problem~$\Pi$ belongs to GLO and $n$ is sufficiently large,
then in view of Proposition~\ref{optimal_bound}, for any ${x} \in
{\rm \mbox{\rm Sol}}$ and ${y} \in \mathcal{N}({x})$, the bitwise
mutation operator with~$p_{\rm m}=K/n$ satisfies the condition
$\Pr\{\mbox{Mut}^*({x})={y}\}^{-1} \in {\rm Poly}$. Therefore,
Corollary~\ref{cor:general} implies the following

\begin{corollary}\label{GA_GLO}
If $\Pi\in {\rm GLO}$ and GA with multistart uses
\begin{enumerate}
\item a polynomial-time initialization procedure that produces
a population with at least one feasible solution with
probability~$p_1$ such that $1/p_1\in {\rm Poly}$,
\item the tournament selection or the $(\mu,\lambda)$-selection,
\item a crossover operator satisfying~(\ref{eps_cross})
or~(\ref{eps_cross1}) for some positive constant~${\varepsilon_0}$
and
\item the bitwise mutation,
\end{enumerate}
then given suitable values of parameters $\lambda, p_{\rm m}$ and
$k$ or $\mu$, GA with multistart visits a solution with a constant
guaranteed approximation ratio within expected polynomially
bounded time.
\end{corollary}

\section{CONCLUSIONS}

The obtained bounds on the first hitting times for sets of global
or local optima are extending some previously known bounds of such
kind for genetic algorithms and may be applied to standard
benchmarks and genetic algorithms as well as some stat-of-the-art
genetic algorithms for combinatorial optimization problems.
Considering the selection operators with very high selection
pressure, we obtain the bounds that apply even in the cases where
the probability of non-downgrading is not lower-bounded by a
positive constant.

The obtained results imply that if a problem is polynomially
bounded and the feasible solutions are present in the initial
population, then a local optimum in a Hamming neighborhood system
is computable in expected polynomial time by standard GAs with
multistart. Besides that, given suitable parameters and
initialization procedure, a non-elitist GA with tournament
selection or $(\mu,\lambda)$-selection approximates any problem
from GLO class within a constant ratio in polynomial time in
expectation.

If an NP optimization problem is polynomially bounded then
Canonical Genetic Algorithm with appropriate parameters tuning and
fitness scaling finds a local optimum within expected polynomial
time for many standard neighborhood systems.



\section*{ACKNOWLEDGEMENT}
The research is supported by RFBR grant~15-01-00785. The author
thanks the participants of the Theory of Evolutionary Algorithms
seminar~15211 at Schloss Dagstuhl for helpful comments.
\\~\\


\end{document}